\pdfoutput=1
\documentclass[11pt]{article}
\usepackage{amsmath}
\usepackage{caption}
\usepackage{acl}

\usepackage{amssymb}
\usepackage{times}
\usepackage{latexsym}
\usepackage{subcaption}
\usepackage{threeparttable}
\usepackage[T1]{fontenc}
\usepackage{multirow}
\usepackage[utf8]{inputenc}
\usepackage{graphicx}
\usepackage{microtype}

\usepackage{inconsolata}

\title{AMA-LSTM: Pioneering Robust and Fair Financial Audio Analysis for Stock Volatility Prediction}

\author{\textbf{Shengkun Wang}$^{1}$, \textbf{Taoran Ji}$^{2}$, \textbf{Jianfeng He}$^1$, \textbf{Mariam Almutairi}$^1$,\\
\textbf{Dan Wang}$^3$, \textbf{Linhan Wang}$^1$, \textbf{Min Zhang}$^1$,\textbf{Chang-Tien Lu}$^1$ \\
$^1${Department of Computer Science, Virginia Tech} \\
$^2${Department of Computer Science, Texas A\&M University - Corpus Christi} \\
$^3${School of Business, Stevens Institute of Technology} \\
}

\begin{document}
\maketitle
\begin{abstract}
Stock volatility prediction is an important task in the financial industry. Recent advancements in multimodal methodologies, which integrate both textual and auditory data, have demonstrated significant improvements in this domain, such as earnings calls\footnote{Earnings calls are public available and often involve the management team of a public company and interested parties to discuss the company's earnings.}. However, these multimodal methods have faced two drawbacks. First, they often fail to yield reliable models and overfit the data due to their absorption of stochastic information from the stock market. Moreover, using multimodal models to predict stock volatility suffers from gender bias and lacks an efficient way to eliminate such bias. To address these aforementioned problems, we use adversarial training to generate perturbations that simulate the inherent stochasticity and bias, by creating areas resistant to random information around the input space to improve model robustness and fairness.
Our comprehensive experiments on two real-world financial audio datasets reveal that this method exceeds the performance of current state-of-the-art solution. This confirms the value of adversarial training in reducing stochasticity and bias for stock volatility prediction tasks.

% Recently, multimodal approaches using both verbal transcript and vocal cues from financial disclosures have outperformed the single-modal methods.

\end{abstract}

\section{Introduction}

In the stock market, predicting the exact price of a stock is deemed impossible \cite{nguyen2015sentiment}, yet it is widely accepted within the financial industry that one can forecast a stock's volatility level using publicly available information \cite{dumas2009equilibrium}. Stock price volatility, defined as the standard deviation of a stock’s returns over a specific period, serves as a commonly used indicator of financial risk. In the past, research efforts have focused on employing time-series models, 
\begin{figure}[!htb]
  \centering
  \begin{subfigure}{.5\columnwidth}
    \centering
    \includegraphics[width=\linewidth]{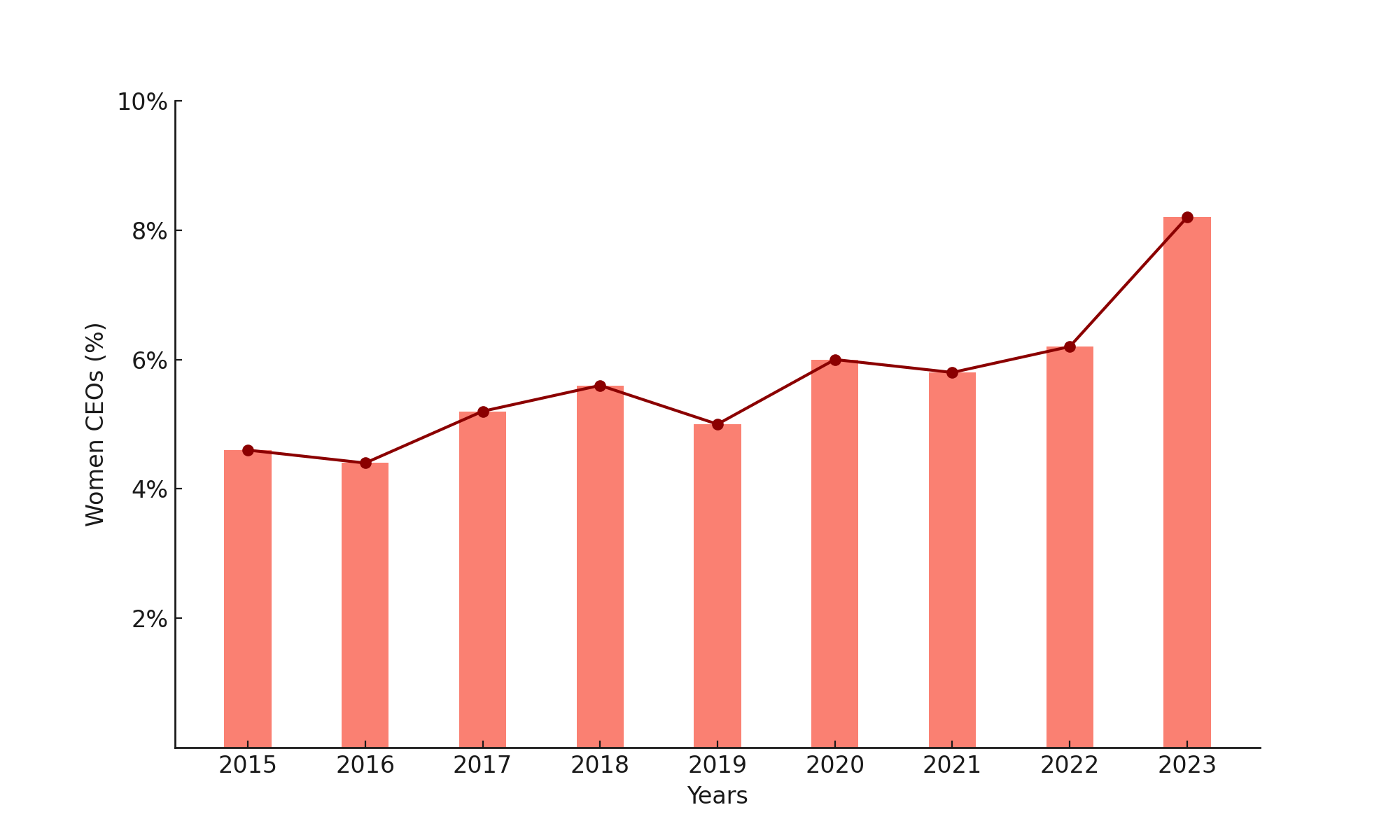}
    \caption{\small S\&P 500 Women CEO}
    \label{fig:sub1}
  \end{subfigure}%
  \begin{subfigure}{.5\columnwidth}
    \centering
    \includegraphics[width=\linewidth]{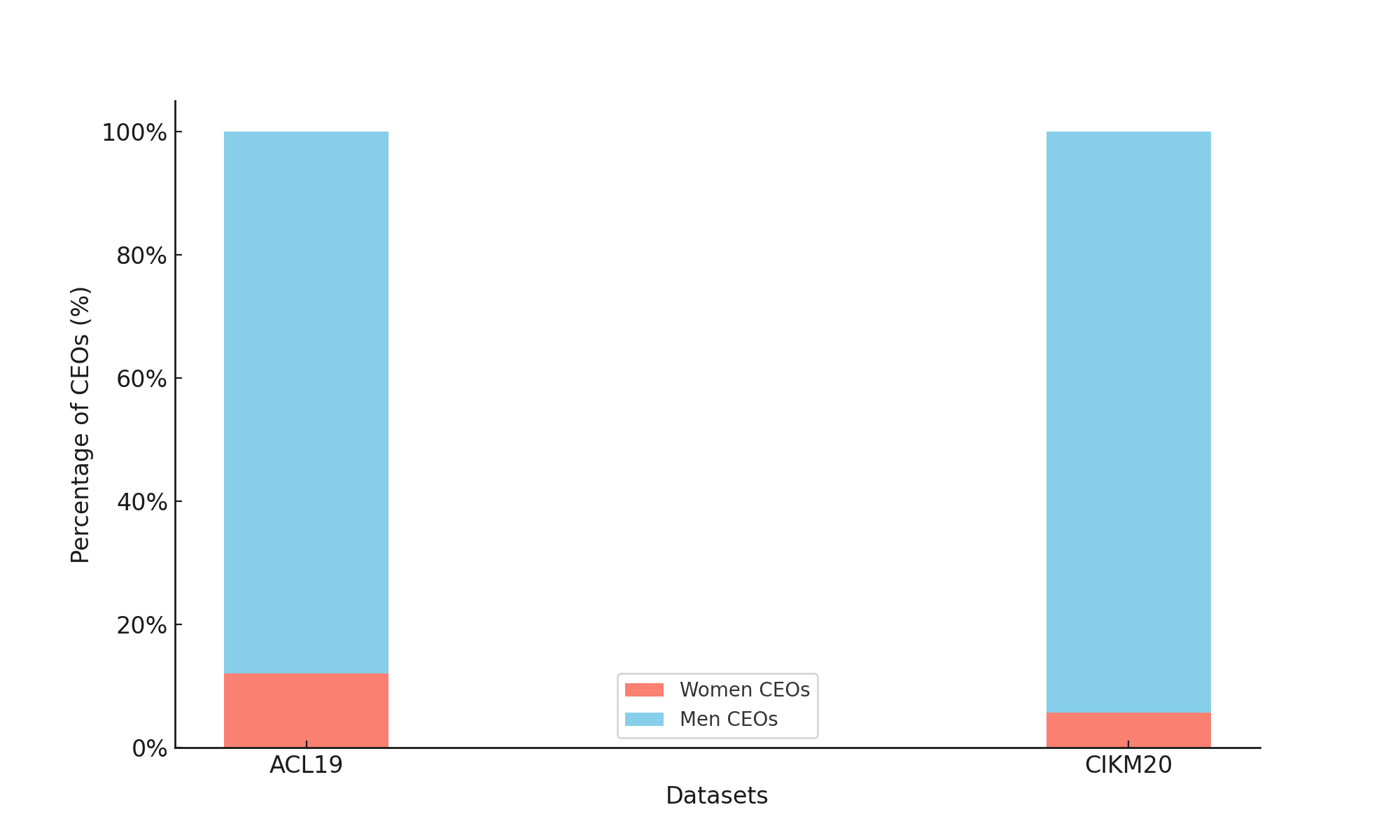}
    \caption{CEO Gender of Datasets}
    \label{fig:sub2}
  \end{subfigure}
  \caption{(a) displays the percentage of woman CEO in recent years, and (b) compares the proportion of female to male CEO within the two datasets utilized.}
  \label{fig:combined}
\end{figure}
using historical stock prices, to predict future volatility \cite{kristjanpoller2014volatility}. In recent years, advancements in Natural Language Processing (NLP) and Speaker Recognition (SR) have opened up new possibilities for the task. For example, researchers leverage novel sources of textual and audio data, including financial news \cite{yang2018explainable}, financial reports \cite{rekabsaz2017volatility}, social media \cite{wang2023stock}, earnings call  \cite{qin2019you} and merger and acquisition (M\&A) call \cite{sawhney2021multimodal} to predict stock volatility. These diverse data sources provide a richer, more nuanced understanding of market dynamics and investor sentiment, significantly enhancing the prediction of stock volatility.

However, the inherent nature of financial data presents two unique challenges. First, textual data is discrete and stochastic \cite{yuan2021bridge}, and audio signals possess a high temporal resolution \cite{donahue2018adversarial}. 
This characteristic makes both text and audio vulnerable to adversarial examples, which can effortlessly mislead human evaluators \cite{carlini2018audio, xie-etal-2022-word}. 
Second, the financial data is biased. Research \cite{das2021fairness} has delved into various biases within this field, with gender imbalance being a particularly prominent issue. \citet{bigelow2014skirting} found that female CEOs, despite having comparable credibility, are perceived as less capable of attracting capital. In the context of earnings calls, \citet{comprix2022role} observed a significant underrepresentation of female executives. This biased financial audio data is further exacerbated by deep learning models, which tend to amplify variations in audio features due to the scarcity of female training examples, leading to gender bias \cite{sawhney2021empirical}.

To address the aforementioned issues, we use adversarial training to enhance the robustness and fairness of financial data interpretation models. Adversarial training was initially used in computer vision tasks \cite{goodfellow2014explaining}, and then expanded to NLP \cite{miyato2016adversarial} and SR \cite{sun2018training} fields. Our method differs from the feature space adversarial training \cite{ijcai2019p810}. It enhances model robustness and reduces output bias by introducing adversarial examples directly from input embeddings, making the approach better aligned with the privacy and proprietary constraints of financial modeling.

We have applied our adversarial training method  to a multimodal attentive LSTM model, which efficiently processes information from two modalities. By introducing perturbations into the input space and dynamically optimizing these perturbations to maximize their impact on the model’s output, we enhance the model's ability to perform well on both clean and perturbed data.
The adversarial training method effectively guarantees stable training and robust performance of the model when dealing with stochastic and biased financial data.
This training approach can also apply to the Transformer, but we've found the attentive LSTM to be more effective for our scenario. This is because the Transformer struggles with temporal information and lacks enough data to learn its many parameters effectively.
To the best of our knowledge, this work is the first one to explore the potential of adversarial training in multimodal learning financial audio task. The main contributions of this paper are summarized as:

% \textbf{\textbullet\ 
% Test.} 

\textbullet\ We suggest a method of adversarial training that tackles the issues of randomness and bias, and implement it on a deep learning model for stock volatility prediction.

\textbullet\ 
We study the impact of gender bias in earnings call audio on stock volatility predictions by training on gender-specific audio features.

\textbullet\ We delve into the stochastic nature of financial data, emphasizing the critical need to address the randomness inherent in input features.

\textbullet\ We test our method on two public benchmarks, showing it outperforms strong baselines and proving that adversarial learning increases its robustness and fairness.

\section{Related Work}
Our work is closely related with the following three lines of research:

\textbf{Stock Volatility with Multimedia Data :}
 Traditional stock volatility prediction method relies on historical pricing data and typically employs both continuous and discrete time-series models \cite{idrees2019prediction}. In addition to stock prices, financial news, analyst reports, earnings reports, and social media have been proven to significantly enhance stock prediction tasks  \cite{wang2023alerta, wang2023stock, zhang2021predicting, wang2020image,zhang2018improving, rekabsaz2017volatility}. Furthermore, recent multimodal models that employ LSTM \cite{qin2019you}, GCN \cite{sawhney2020voltage} and Transformer \cite{yang2020html}  to extract features from earnings calls audio and combine it with textual transcript, have achieved state-of-the-art results in stock volatility prediction.

\textbf{Bias in Financial Audio Data :}
Financial audio features can indicate a speaker's emotional and psychological state \cite{fish2017sound}. Previous research has demonstrated that audio features, such as pitch and intensity, differ significantly across genders \cite{burris2014quantitative}. Especially in earnings calls, female executives are highly underrepresented,  \citet{suresh2019framework} noted that only 5\% of Fortune-500 CEOs are women. Moreover, under identical conditions, men are often perceived as more charismatic than female executives \cite{novak2017gender}. This disparity in audio features becomes more pronounced in deep learning models due to the scarcity of female training examples, leading to the manifestation of gender bias.

% In the financial domain, while many studies have utilized historical pricing data, textual, and tabular data, we focus on learning from both text and audio data \cite{yang2020html}. Recent advances in deep learning have significantly improved multi-modal learning tasks, where high-level embeddings from different data types are integrated via deep neural networks \cite{ma2019exploring}.
% Moreover, the interplay between text and audio data in a multi-modal learning framework has been a topic of interest in speech communication research.
% Studies have found that acoustic features are closely linked with emotions \cite{haq2009speaker}, trustworthiness \cite{belin2017sound}, and confidence \cite{jiang2017sound}.
% Despite these insights, the application of audio data in financial contexts is still at an early stage \cite{mathur2022docfin}.
% % Given the effectiveness of recent multi-modal approaches, and the availability of task-relevant text and audio data for volatility forecasting, it is clear that these techniques warrant further consideration, hence the approach is taken in the present work.
\begin{figure*}[h]
    \centering
    \includegraphics[width=\textwidth]{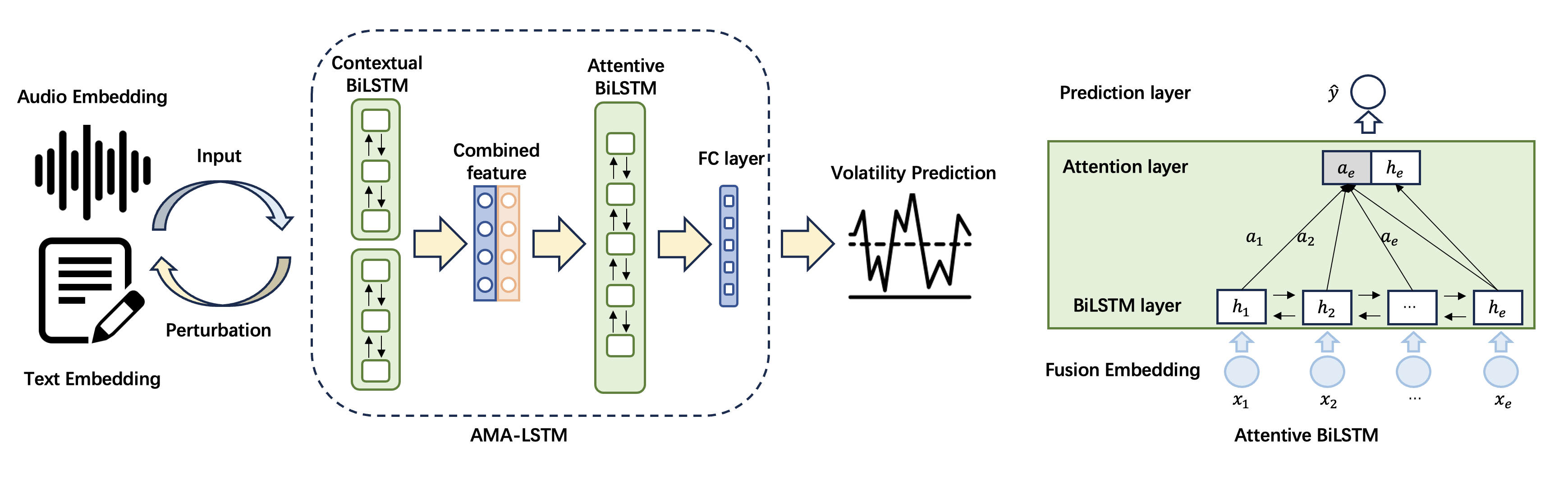}
    \caption{Illustration of the adversarial multimodal attentive LSTM architecture and an attentive BiLSTM block. }
    \label{fig:p2}
\end{figure*}

\textbf{Adversarial Training :}
The concept of adversarial training is quite direct, it enhances training data by incorporating adversarial examples in each iteration of the training process. Initially brought to the forefront by \citet{szegedy2013intriguing}, this idea involved training neural networks with a combination of adversarial and clean examples \cite{goodfellow2014explaining}.
% expanded on this by introducing the Fast Gradient Sign Method (FGSM) to generate adversarial examples during training.
\citet{huang2015learning} further developed this concept by framing adversarial training as a min-max optimization problem.
Adversarial training is now widely recognized as one of the most effective methods for boosting the robustness of deep learning models  \cite{athalye2018robustness}.
The existing works mainly concentrate on computer vision tasks \cite{goodfellow2014explaining}. Due to the simplicity and effectiveness, adversarial training have recently extended its application to NLP \cite{zang2019word} and SR \cite{mei2022diverse}. In these areas, researchers generate adversarial examples from input embeddings, such as word and audio embeddings, to create perturbation-resistant areas around the input space.

% However, adversarial training, in the feature space is still in an early stage. \cite{ijcai2019p810} proposes a 
% Although this work is inspired by these adversarial learning research efforts, it targets a distinct task—stock volatility prediction, of which the data are time series with stochastic property. To the best of our knowledge, this work is the first one to explore the potential of adversarial training in time-series analytics \cite{sawhney2021empirical}.

\section{Problem Formulation}
The stock volatility prediction problem is formulated following \cite{kogan2009predicting}. For a given stock with an adjusted closing price \( p_i \) on trading day \( i \), we calculate the average log volatility over \( n \) days following the earnings call as follows:

\begin{equation}
\upsilon_{[0,n]} = \ln \left( \sqrt{\frac{\sum_{i=1}^{n} (r_i - \bar{r})^2}{n}} \right),
\end{equation}

\noindent where \( r_i \) is the stock return on day \( i \) and,  \( \bar{r} \) represents the average stock return across a \( n \) days window. We define the return as \( r_i = \frac{p_i }{p_{i-1}} - 1 \).  Based on the value of n, the horizon of the prediction can be adjusted. Here, we consider various window sizes n,  such as 3, 7, and 15 trading days.
In our study, we have \( P \) earnings call transcripts and the longest one has \( Q \) sentences. We denote \( x = \{x_1, \ldots, x_e\}\) as the multimodal fusion embedding and \( x_e \) is the fusion embedding for \( e\text{-}th \) earnings call where \( x_e \ \in \mathbb{R}^{Q \times D} \). This embedding \( x_e \) encompasses an audio component \( A_e \in \mathbb{R}^{Q \times D_a} \), corresponding aligned text component \( T_e \in \mathbb{R}^{Q \times D_t} \), where \(D_a \) and \(D_t \) denote the dimensions of audio and text embeddings, respectively. The total dimension \(D\) is the concatenation of these two. 
Our goal is to develop a predictive regression function \( f(x_e) \rightarrow \upsilon_{[0,n]} \).

\section{Adversarial Multimodal Attentive LSTM Architecture }
Our adversarial multimodal attentive LSTM (AMA-LSTM)  has two parts. We first introduce the architecture of the attentive multimodal model, which operates without adversarial training. Then, we specify the attentive LSTM block, which contains four components: feature mapping layer, LSTM layer, attention layer, and prediction layer. Secondly, we thoroughly explain how adversarial training helps multimodal models improve robustness and fairness.

\subsection{Multimodal Attentive LSTM}
The attentive multimodal model comprises two primary components. As shown in Figure 2, the first two contextual LSTM blocks are designed to extract unimodal features from either text or audio data.  These blocks adeptly capture relationships and dependencies within each individual modality.  In the second component, the extracted features from both text and audio modalities are combined and fed into an attentive LSTM block, followed by a fully-connected layer.

\noindent\textbf{LSTM layer}. 
LSTM's \cite{hochreiter1997long} proficiency in capturing long-term dependencies has made it a popular choice for processing multimodal sequential data \cite{qin2019you}. The core principle of LSTM involves recurrently transforming an input sequence into a series of hidden representations. At each time-step, LSTM updates the hidden representation \( h_e \) by integrating the current input \( x_e \) with the preceding hidden representation \( h_{e-1} \), thereby capturing the sequential dependencies : \( h_e = LSTM(x_e, h_{e-1}) \). Adapting this concept, we employ a BiLSTM layer to better capture the bidirectional temporal patterns and sequential dependencies in text and audio features. The layer maps the sequence 
\( [x_1, \ldots, x_e] \) into hidden representations  \( [h_1, \ldots, h_e] \in \mathbb{R}^{U \times D} \) with the dimension of \(U \).

\noindent\textbf{Attention Layer}.  
The attention mechanism compresses hidden representations from different time-steps into a unified overall representation and assigns adaptive weights to each step. The key concept here is the recognition that data from different time-steps may vary in their contribution to the overall sequence representation. For financial data representation, status at different time-steps might also contribute differently. As such, we use an attention mechanism to aggregate the hidden representations as:
\vspace{-1em}

\begin{equation}
\begin{gathered}
a_e = \sum_{e=1}^{E} \alpha_e h_e, 
\alpha_e = \frac{\exp{(c_e)}}{\sum_{e=1}^{E} \exp{(c_e)}}, \\
\alpha_e = u \tanh(\mathbf{W}_a \mathbf{h}_e + \mathbf{b}_a),
\end{gathered}
\end{equation}
where \( W_a \in \mathbb{R}^{K \times U} \), \( b_a \) and \( u \in \mathbb{R}^{K} \) are parameters to be learned.

\noindent\textbf{Prediction Layer}. 
Instead of predicting directly from \( a_e \), we first involves concatenating \( a_e \) with the last hidden state \( h_e \) to form the final latent representation of the earnings call:

\begin{equation}
l_e = [a_e, h_e],
\end{equation}

\noindent where \( l_e \in \mathbb{R}^{2U} \). The rationale behind this is to give additional emphasis to the most recent time-step, which is often considered highly indicative of future volatility. Utilizing \( l_e \), we then apply a fully connected layer as our predictive function. This layer is responsible for estimating the classification confidence \( y_e \),  formulated as \( {y}_e = w_m l_e + b_m \).

\subsection{Adversarial Training}
Applying multimodal attentive  LSTM models to forecast stock volatility presents inherent challenges due to the stochastic and biased nature of financial data, notably from earnings calls \cite{sawhney2021empirical}. Earnings calls are rich with qualitative information that is often speculative and sentiment-driven, contributing to the stochastic and biased nature of the data features  \cite{blau2015sophisticated}. An adversarial training approach counters this by perturbing input data to simulate these uncertainties, thereby pushing the model to maintain robustness predictions and reduce bias. This method, a strategic deviation from training solely on clean data, aims for robustness by incorporating the worst-case scenarios within its optimization function. The goal is to craft models that are both sensitive to the nuanced dynamics of the market and resistant to overfitting, ensuring reliable performance and robustness to the financial task.

\begin{figure}[h]
    \centering
    \includegraphics[width=\columnwidth]{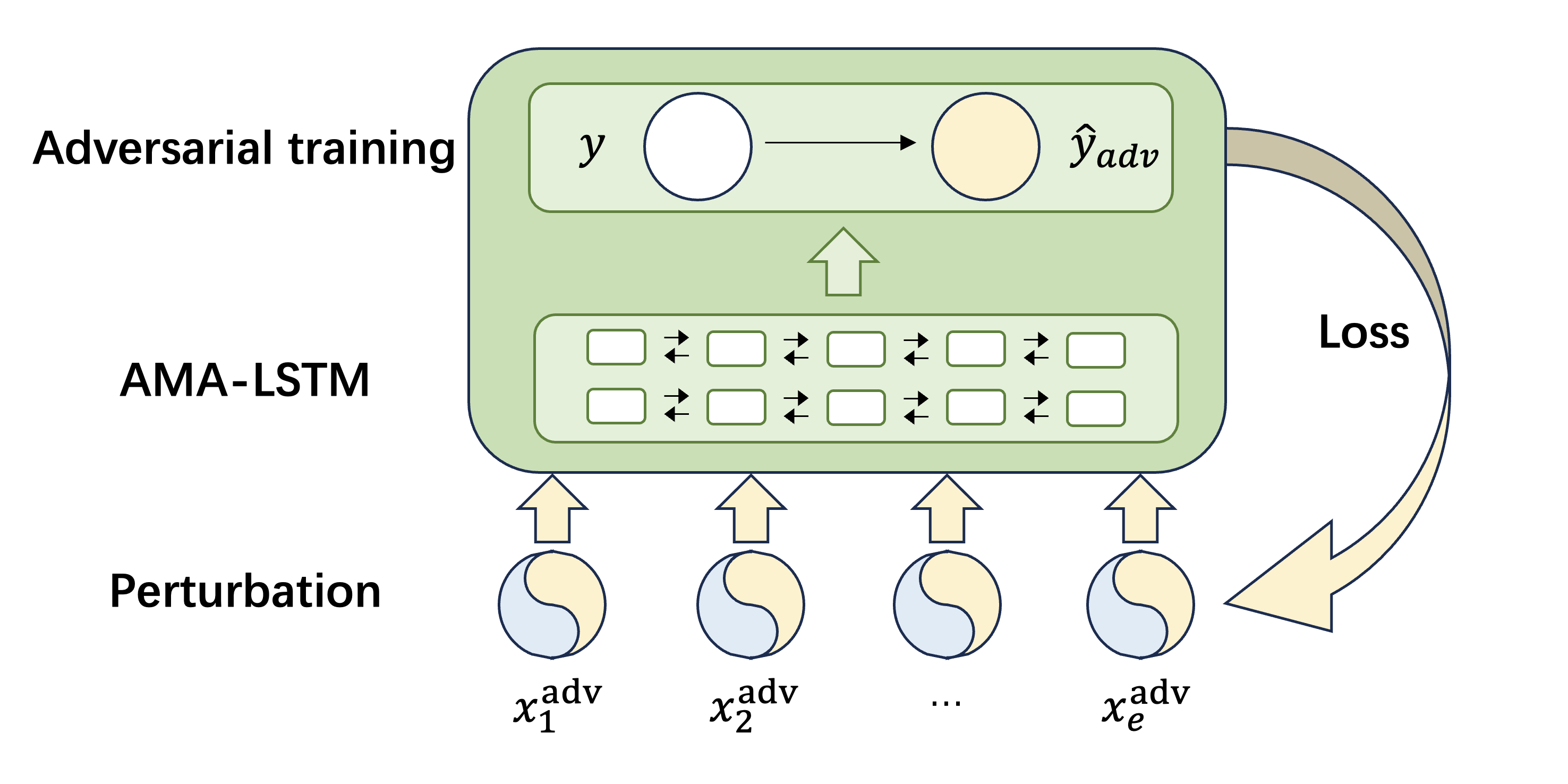}
    \caption{Illustration of the AMA-LSTM adversarial training process. Perturbations (\(\delta\)) are derived by computing the gradients of the token embeddings in relation to the loss function.}
    \label{fig:p3}
\end{figure}

As shown in Figure 3, in developing an adversarial training model for financial data, inspiring by \citet{shaham2018understanding} we approach the problem through robust optimization.  The adversarial training process seeks to solve the objective function of AMA-LSTM:

\begin{equation}
\min_{\theta} \mathbb{E}_{(x,y)\sim D} \left[ \max_{\delta \in S} \left( f_{\theta}(x + \delta) - y \right)^2 \right] + \lambda R(\theta)
,
\end{equation}

\noindent where \( (x,y) \sim \mathcal{D} \)  denotes the training data drawn from the distribution \( \mathcal{D} \), 
 $S$ defines the allowed range of perturbations and \(R(\theta)\) represents the L2 regularization term, which is the sum of the squared norms of the parameters, expressed as \(R(\theta) = \sum_i \|\theta_i\|^2.\) 
This method frames adversarial training as a min-max problem, focusing on minimizing the regression error while simultaneously considering an adversary's efforts to maximize this error through input perturbations. The essence of this approach is the generation of strong adversarial examples that push the model to find the best possible parameters under worst-case scenarios. Therefore, adversarial training, by creating anti-disturbance areas, could enable the multimodal model to capture the stochastic and biased properties of earnings calls input. Since it is intractable to directly calculate \(\delta\)
we employed a multi-step gradient based attack method \cite{madry2017towards} for solving the problem as follows:

{\small
\begin{equation}
x^{t+1}_{adv} = \text{Proj}_{x+S} \left( x^t_{adv} + \beta \text{ sign}(\nabla_x \left( f_{\theta}(x^t_{adv}) - y \right)^2) \right),
\end{equation}
}

\noindent where t is the current step and \( \beta \) is the step size, it would lead to the largest change on the model prediction significantly increased the adversarial robustness and fairness of deep learning models against a wide range of bias.

% Further, they investigated the inner maximization problem from the landscape of adversarial examples and gave both theoretical and empirical proofs of local maxima's tractability with PGD. 

\begin{table}[ht]
\centering
\caption*{\small Table 1: Performance comparison on the two datasets. MSE of different models on stock volatility prediction n days following the earnings call. }
\resizebox{\columnwidth}{!}{%
\begin{tabular}{cccc|ccc}
\hline
\multirow{2}{*}{Methods} & \multicolumn{3}{c|}{ACL19} & \multicolumn{3}{c}{CIKM20} \\ \cline{2-7} 
                        & $\text{MSE}_3$    & $\text{MSE}_7$   & $\text{MSE}_{15}$   & $\text{MSE}_3$    & $\text{MSE}_7$   & $\text{MSE}_{15}$   \\ \hline
bc-LSTM                 & 1.41      & 0.44     & 0.30      & 1.44      & 0.51     & 0.35      \\
MDRM                    & 1.37      & 0.42     & 0.30      & 1.43      & 0.48     & 0.32      \\
HTML                    & 0.85      & \textbf{0.35}     & 0.25      & 1.15      & 0.47     & 0.30      \\
M3ANet                  & 0.72      & 0.38     & 0.24      & 0.86      & 0.41     & 0.31      \\
AMA-LSTM                & \textbf{0.68}      & 0.36     & \textbf{0.23}      & \textbf{0.74}      & \textbf{0.35}     & \textbf{0.27}      \\ \hline
\end{tabular}%
}
\label{t1}
\end{table}

\begin{table}[ht]
\centering
\caption*{\small Table 2: \(\Delta\)MSE is the difference in MSE between female and male CEO distributions over 3, 7, and 15 days. Our method shows the
best performance.} 
\resizebox{\columnwidth}{!}{%
\begin{tabular}{cccc|ccc}
\hline
\multirow{2}{*}{Methods} & \multicolumn{3}{c|}{ACL19} & \multicolumn{3}{c}{CIKM20} \\ \cline{2-7} 
                        & \small\(\Delta\)$\text{MSE}_3$    & \small\(\Delta\)$\text{MSE}_7$   & \small\(\Delta\)$\text{MSE}_{15}$   & \small\(\Delta\)$\text{MSE}_3$    & \small\(\Delta\)$\text{MSE}_7$   & \small\(\Delta\)$\text{MSE}_{15}$   \\ \hline
bc-LSTM                 & 0.38      & 0.23     & 0.27      & 0.42      & 0.31     & 0.24      \\
MDRM                    & 0.30      & 0.11     & 0.28      & 0.36      & 0.33     & 0.29      \\
HTML                    & 0.33      & 0.14     & 0.28      & 0.29      & 0.25     & 0.16      \\
M3ANet                  & 0.24     & 0.10     & 0.26     & 0.31      & 0.28     & 0.20      \\
AMA-LSTM                & \textbf{0.19}      & \textbf{0.08}     & \textbf{0.15}      & \textbf{0.23}      & \textbf{0.21}     & \textbf{0.13}      \\ \hline
\end{tabular}%
}
\label{t2}
\end{table}

\section{Experiments}
We conducted experiments to answer the following questions
about the performance of AMA-LSTM: 

\noindent Q1. \textbf{Robustness}. 
Since earnings calls often contain much extraneous information unrelated to company performance, whether the adversarial training reduce stochastic of the financial data.

\noindent Q2. \textbf{Fairness}. Due to the differences in verbal and vocal cues across genders, predictions made using gender-specific data may yield varied results. The question arises whether adversarial training can reduce the bias originating from these differences.

\noindent Q3. \textbf{Ablation study}. Between text and audio information, which exhibits greater bias.

\subsection{Dataset}
We utilized two real-world datasets from the financial industry: ACL19 \cite{qin2019you} contains 576 public earnings calls audio recordings with their transcripts for 277 companies in the S\&P 500. CIKM20 \cite{li2020maec}  comprise 3443 earnings call embeddings along with transcripts for 1213 companies in the S\&P 1500. Additionally, we sourced the necessary dividend-adjusted closing prices for volatility prediction via the Yahoo Finance \footnote{\url{https://finance.yahoo.com/}} API by using the relevant stock tickers.
Besides, due to various corporate changes such as mergers, acquisitions, and rebranding, several companies have changed their names and stock tickers. We manually collected information regarding these stocks from Investing\footnote{\url{https://www.investing.com/}}. In terms of data processing, we followed the approach outlined by \cite{sawhney2021empirical} for textual features, employing FinBERT embeddings with their default settings \footnote{\url{https://github.com/ProsusAI/finBERT}}\cite{araci2019finbert}. For audio cues, we extracted multi-dimensional feature vectors using Praat \cite{boersma2001speak}.

\subsection{Evaluation metrics}
Following \cite{qin2019you,sawhney2021empirical}, we evaluate the accuracy of our volatility predictions by comparing the predicted values \( y_i \) with the actual volatility values   \( \hat{y}_i \). We use the Mean Squared Error (MSE) as our evaluation metric, defined as:

\begin{equation} \label{eq:mse}
MSE = \frac{\sum_{i} (y_i - \hat{y}_i)^2}{n} .
\end{equation}

Furthermore, to evaluate gender bias in our model, we measure the disparity in performance errors between genders, denoted as \(\Delta MSE = MSE_f - MSE_m\), where \(f\) represents female \(m\) represents male. A higher value of \(\Delta MSE\) indicates a bias favoring male-oriented data.

\subsection{Baselines}
In this section, we compare four baselines. These methods represent previous approaches to stock volatility prediction that utilized either LSTM or transformer-based multimodal models.

\noindent\textbullet\textbf{bc-LSTM} \cite{poria2017context} : employs separate contextual Bi-LSTMs to extract uni-modal features then fused together.

\noindent\textbullet\textbf{MDRM} \cite{qin2019you} : utilizes pretrained GloVe embeddings and hand-crafted acoustic features. These are processed through separate BiLSTMs to obtain their uni-modal contextual embeddings. Then, these embeddings are fused and input into a two-layer dense network.

\noindent\textbullet\textbf{HTML} \cite{yang2020html} : employs WWM-BERT to encode text tokens. It then fuses the unimodal features and inputs them into a sentence-level transformer.

\noindent\textbullet\textbf{M3ANet} \cite{sawhney2021multimodal} : utilizes uncased base BERT to encode text tokens. It attentively fuses the unimodal features and then inputs them into a sentence-level transformer.

\noindent\textbf{Performance Comparison.} Table 1 displays the prediction performance, measured by MSE, of the compared methods on two datasets. And Table 2 displays the differences in prediction accuracy between earnings calls led by female and male CEOs. The result leads to the following observations:

\begin{itemize}
    \item Our method achieves the best MSE in almost all cases. It  surpasses the previous state-of-the-art Transformer-based methods and LSTM-based approaches. Specifically, AMA-LSTM beats Transformer-based M3ANet by 20.83\% and 13.95\% across two datasets. These results indicate that through simulated perturbations during adversarial training, the model creates areas resistant to randomness information around the input space. This reduces the interference of stochasticity in financial data on the model's predictions, thereby enhancing the model's robustness.
\end{itemize}

\begin{itemize}
    \item We evaluate the gender bias by comparing \(\Delta\)MSE, showing that AMA-LSTM achieves the best results in both short-term and long-term across two datasets compared to models without adversarial training. These results demonstrate that adversarial training diminishes the model's sensitivity to gender-specific features, thereby improving the fairness. It also highlights the need for increased attention to bias issues within financial data.

\end{itemize}

\begin{table}[]
\centering
\caption*{\small Table 3: Performance of stochastic perturbation on the two datasets}
\resizebox{0.6\columnwidth}{!}{%
\begin{tabular}{cccc}
\hline
Datasets & $\text{MSE}_3$ & $\text{MSE}_7$ & $\text{MSE}_{15}$ \\ \hline
ACL19    & 0.81   & 0.38   & 0.28    \\
CIKM20   & 0.97   & 0.48   & 0.32    \\ \hline
\end{tabular}%
}
\label{t3}
\end{table}

\noindent\textbf{Stochastic perturbation.}
We also show the robustness of adversarial training by comparing its effectiveness against both adversarial and random perturbations. 
Stochastic multimodal attentive LSTM (SMA-LSTM) is a variation of AMA-LSTM. It creates additional examples by introducing random perturbations to clean input examples. Table 3 displays SMA-LSTM's performance across two datasets. A comparison with Table 3 reveals that: 1) AMA-LSTM significantly outperforms random perturbation. 
Specifically, adversarial training outperforms stochastic perturbation by 16.05\% on the ACL19 and 23.71\% on the CIKM. 
\begin{table}[ht]
\centering
\caption*{\small Table 4: An ablation study was conducted on AMA-LSTM. In this context, AMA-LSTM(A) represents the model using only audio, AMA-LSTM(T) stands for model using only text, and the final variant incorporates both audio and text. }
\resizebox{\columnwidth}{!}{%
\begin{tabular}{cccc|ccc}
\hline
\multirow{2}{*}{Methods} & \multicolumn{3}{c|}{ACL19} & \multicolumn{3}{c}{CIKM20} \\ \cline{2-7} 
                        & \small\(\Delta\)$\text{MSE}_3$    & \small\(\Delta\)$\text{MSE}_7$   & \small\(\Delta\)$\text{MSE}_{15}$   & \small\(\Delta\)$\text{MSE}_3$    & \small\(\Delta\)$\text{MSE}_7$   & \small\(\Delta\)$\text{MSE}_{15}$   \\ \hline
AMA-LSTM (A)                    & 0.25      &  0.14    & 0.23      & 0.27      & 0.26     & 0.17      \\
AMA-LSTM (T)                  & 0.22      & 0.13     & 0.23      & 0.24      & 0.23     & 0.14      \\
AMA-LSTM                & 0.19     & 0.08     & 0.15      & 0.23      & 0.21     & 0.13      \\ \hline
\end{tabular}%
}
\label{t1}
\end{table}
This shows that adversarial perturbations can improve stock volatility prediction by enhancing model robustness. 2) SMA-LSTM exceeds the performance of the LSTM baseline, which is trained only on clean examples. This emphasizes the importance of addressing the stochastic nature of financial data.

\subsection{Ablation Study}
An ablation study we have constructed two variations alongside the fully-loaded model. Each variant is specifically tailored to handle certain types of input data: AMA-LSTM(A) solely relies on audio data, AMA-LSTM(T) exclusively processes text embedding. As shown in Table 4, we find that audio data contains more bias than textual information.  This can be attributed to audio features differ greatly between males and females, and the uneven gender distribution among speakers in earnings calls amplifies this discrepancy.
This result indicates that when processing financial data, there should be a greater focus on the bias in audio data. Furthermore, by merging data from various
modalities, the model shows reduced sensitivity to bias and enhanced robustness compared to relying on just one modality.

\section{Conclusion}
Our research has shown that neural networks used for predicting stock market volatility often face challenges in robustness, particularly in handling the stochasticity and bias inherent in financial audio data. To tackle this, we introduced an innovative solution: the adversarial multimodal attentive LSTM. This method employs adversarial training to more effectively simulate the market noise and biases during model training. The experiments on two benchmark datasets not only validated the effectiveness of our approach but also underscored the importance of considering the stochasticity and bias in financial data for stock volatility prediction tasks. The results further revealed that adversarial training significantly enhances the robustness and fairness of the prediction models. 
\bibliography{anthology,custom}

\end{document}